\documentclass[conference]{IEEEtran}
\IEEEoverridecommandlockouts
\usepackage{cite}
\usepackage{amsmath,amssymb,amsfonts}
\usepackage{graphicx}
\usepackage{textcomp}
\usepackage{xcolor}
\def\BibTeX{{\rm B\kern-.05em{\sc i\kern-.025em b}\kern-.08em
    T\kern-.1667em\lower.7ex\hbox{E}\kern-.125emX}}

\usepackage{xurl}
\usepackage{amsthm}%
\usepackage{hyperref}       
\ifCLASSOPTIONcompsoc
    \usepackage[caption=false, font=normalsize, labelfont=sf, textfont=sf]{subfig}
\else
\usepackage[caption=false, font=footnotesize]{subfig}
\fi
\usepackage{booktabs}

\newtheorem{definition}{Definition}

\makeatletter 

\let\old@ps@IEEEtitlepagestyle\ps@IEEEtitlepagestyle
\def\confheader#1{%
    \def\ps@IEEEtitlepagestyle{%
        \old@ps@IEEEtitlepagestyle%
        \def\@oddhead{\strut\hfill#1\hfill\strut}%
        \def\@evenhead{\strut\hfill#1\hfill\strut}%
    }%
    \ps@headings%
}
\makeatother

\confheader{%
        \parbox{20cm}{\textbf{\textit{Accepted at IEEE Conference on Artificial Intelligence (CAI 2024).}}}
}

\begin{document}
\title{Low Variance Off-policy Evaluation with State-based Importance Sampling \thanks{DMB was supported by UKRI Trustworthy Autonomous Systems Hub, EP/V00784X/1 and PST was supported by NSF grant no. CCF-2018372.}}
\author{\IEEEauthorblockN{1\textsuperscript{st} David M. Bossens}
\IEEEauthorblockA{\textit{IHPC}\\
 \textit{Agency for Science, Technology and Research}\\ 
 Singapore\\
\textit{CFAR}\\ 
\textit{Agency for Science, Technology and Research}\\ 
Singapore \\
bossensdm@cfar.a-star.edu.sg}
\and
\IEEEauthorblockN{2\textsuperscript{nd} Philip S. Thomas}
\IEEEauthorblockA{\textit{Manning College of Information and Computer Sciences} \\
\textit{University of Massachusetts}\\
Amherst, United States \\
pthomas@cs.umass.edu}
}

\maketitle

\begin{abstract}
In many domains, the exploration process of reinforcement learning will be too costly as it requires trying out suboptimal policies, resulting in a need for off-policy evaluation, in which a target policy is evaluated based on data collected from a known behaviour policy. In this context, importance sampling estimators provide estimates for the expected return by weighting the trajectory based on the probability ratio of the target policy and the behaviour policy. Unfortunately, such estimators have a high variance and therefore a large mean squared error. This paper proposes state-based importance sampling estimators which reduce the variance by dropping certain states from the computation of the importance weight. To illustrate their applicability, we demonstrate state-based variants of ordinary importance sampling, weighted importance sampling, per-decision importance sampling, incremental importance sampling, doubly robust off-policy evaluation, and stationary density ratio estimation. Experiments in four domains show that state-based methods consistently yield reduced variance and improved accuracy compared to their traditional counterparts.
\end{abstract}


\maketitle

\section{Introduction}
In reinforcement learning \cite{Sutton2018}, an agent explores an environment to learn a policy that optimises a utility function, such as the expected cumulative reward, in that environment. 
In many domains, including medical interventions (e.g. surgery, diabetes or sepsis), financial decision-making (e.g. advertising or trading), and safe 
navigation, it would be preferable to avoid costly trial-and-error and instead use a known expert or safe policy to gather the data needed for reinforcement learning. Such domains motivate the use of off-policy evaluation \cite{Udagawa2022,Levine2020}, a class of methods which  estimate the utility of a target policy based on trajectories from a behaviour policy.

Importance sampling (IS) is a traditional statistical technique which estimates an integral over one distribution of interest while only having 
samples from another distribution. The expected cumulative reward can be seen as such an integral, and therefore IS is an 
important technique for off-policy evaluation \cite{Precup2000,Thomas2015,Chandak2020a}. Such techniques have been used successfully in the context 
of diabetes treatment \cite{Chandak2020a} and digital marketing \cite{Thomas2015}. Unfortunately, the variance of IS based off-policy evaluation grows exponentially with the horizon $H$ of the decision process, a problem also known as ``the curse of horizon'' \cite{Liu2018}. This makes IS based off-policy evaluation perform more poorly in domains requiring long-term planning (e.g. safe navigation).

To reduce the variance of off-policy evaluation, this paper proposes a technique that can be used on a wide variety of estimators. The paper makes the following contributions:
\begin{itemize}
\item state-based importance sampling, a class of state-based importance sampling estimators which remove ``negligible states'' from the importance weight computation;
\item two methods to identify negligible states, one based on covariance testing \cite{Guo2017} and one based on state-action values; 
\item implementations of state-based variants of ordinary importance sampling, weighted importance sampling \cite{Precup2000,Precup2001}, per-decision importance sampling \cite{Precup2000}, incremental importance sampling \cite{Guo2017}, doubly robust off-policy evaluation \cite{Thomas2016}, and stationary density ratio estimation  \cite{Liu2018}; and
\item a set of empirical experiments demonstrating the performance of these state-based estimators compared to their traditional counterparts.
\end{itemize}

\section{Problem statement and preliminaries}
This present paper is based on finite-horizon undiscounted Markov decision processes (MDPs), which can be defined based on the tuple $(\mathcal{S},\mathcal{A},r,\mathcal{T})$, where $\mathcal{S}$ is the state space, $\mathcal{A}$ is the action space, $r: \mathcal{S} \times \mathcal{A} \to \mathbb{R}$ is a reward function, 
and $\mathcal{T}: \mathcal{S} \times \mathcal{A}  \to \Delta^{\mathcal{S}}$ is the transition dynamics model defined over the probability simplex over 
the state space, $\Delta^{\mathcal{S}}$. The goal of the reinforcement learning agent is to select actions that maximise the expected cumulative 
reward, which is also called the expected return. The decision-making is represented by the policy of the agent, which takes the form $\pi: \mathcal{S} \to \Delta^{\mathcal{A}}$; that is, the policy $\pi$ outputs the action distribution for each given state.

Off-policy evaluation computes the expected return $\mathcal{G}$ of an evaluation policy $\pi_{e}$ based on samples from a behaviour policy $\pi_b$. This can be done using importance sampling,
\begin{align*}
\mathcal{G} &= \mathbb{E}_{\pi_e}\left[ \sum_{t=1}^{H} r_t \right]  \\
  &= \mathbb{E}_{\pi_b}\left[ \sum_{t=1}^{H} r_t \left( \prod_{t=1}^{H} \frac{\pi_e(a_t|s_t)}{\pi_b(a_t|s_t)} \right) \right] \,,
\end{align*}
which samples from $\pi_b$ and then corrects for the probability under $\pi_e$, using the product $\prod_{t=1}^{H} \frac{\pi_e(a_t^{(i)}|s_t^{(i)})}{\pi_b(a_t^{(i)}|s_t^{(i)})}$ as the importance weight of the trajectory.

A long-standing problem with importance sampling in this manner is that if the horizon is large, then the variance on the estimate of $\mathcal{G}$ is prohibitively large. To see why, define the estimate $\hat{G}_{IS}$
\begin{align*}
\hat{G}_{IS} &= \frac{1}{n} \sum_{i=1}^{n}  \sum_{t=1}^{H} r_t^{(i)} \left( \prod_{t=1}^{H} \frac{\pi_e(a_t^{(i)}|s_t^{(i)})}{\pi_b(a_t^{(i)}|s_t^{(i)})} \right) \,,
\end{align*}
where $i$ indexes the trajectory of the state $s_t^{(i)}$, action $a_t^{(i)}$, and reward $r_t^{(i)}$ at time $t$.
By sample mean of a random variable $X = \sum_{t=1}^{H} r_t \left( \prod_{t=1}^{H} \frac{\pi_e(a_t|s_t)}{\pi_b(a_t|s_t)} \right)$ and Popoviciu's inequality \cite{Popovicu1935},
\begin{align}
\label{eq: var ordinary}
\text{Var}(\hat{G}_{IS}) &= \frac{1}{n} \text{Var}(X) \nonumber \\
					&\leq \frac{1}{4n} \left(\sum_{t=1}^{H} r_{\text{max}} \left( \prod_{t=1}^{H} \rho_{\text{max}} \right)  \right)^2 \nonumber   \\
					&= \frac{1}{4n} \left(H r_{\text{max}} \rho_{\text{max}}^H\right)^2 \nonumber \\
					&= \mathcal{O}(\exp(H)) \,,
\end{align}
where $\rho_{\text{max}} = \max_{(s,a) \in \mathcal{S} \times \mathcal{A}} \pi_e(a|s)/\pi_b(a|s)$ and $[0,r_{\text{max}}]$ is the reward range. In addition to the general upper bound, it is also possible, with some additional assumptions, to show that the variance is lower bounded by $\Omega(\exp(H))$ \cite{Liu2020a}. Due to the bias-variance decomposition of the mean squared error,
\begin{align*}
\text{MSE}(\hat{G}_{IS}) = \text{Bias}(\hat{G}_{IS})^2 + \text{Var}(\hat{G}_{IS}) \,,
\end{align*}
where $\text{Bias}(\hat{G}_{IS}) = \mathbb{E}_{\pi_b}[\hat{G}_{IS}] - \mathcal{G}$, the above results imply an exponential dependency of the MSE on the horizon of the decision problem.

\section{Related work}
Due to the curse of horizon, variance reduction is one of the important problems in off-policy evaluation. We review variance reduction techniques and provide exemplary off-policy evaluation applications.
\subsection{Variance reduction in importance sampling}
While ordinary importance sampling takes a simple average across trajectories, dividing by a denominator $n$, the weighted importance sampling (WIS) technique \cite{Precup2000,Precup2001} divides by the sum of the importance weights of the different trajectories. This mitigates the impact of excessively large importance ratios. Its resulting variance converges to 0 for $n \to \infty$ \cite{Precup2001} but remains exponential in the horizon \cite{Liu2018}. While state-based importance sampling does not work similarly to weighted importance sampling, it shares the commonality that it serves as an easy plug-in to a wide range of estimators.

Another common technique is to remove some of the time steps from the cumulative product, which results in a lower exponent in Eq.~\ref{eq: var ordinary} and mitigates the curse of horizon. In per-decision importance sampling (PDIS) \cite{Precup2000}, one forms different importance weights for each reward $r_{t'}$ by relating it to the past time steps until that point, i.e. $\rho_{t'} = \prod_{t=1}^{t'} \frac{\pi_e(a_t^{(i)}|s_t^{(i)})}{\pi_b(a_t^{(i)}|s_t^{(i)})}$. Incremental importance sampling (INCRIS) \cite{Guo2017} drops all but the $k$ most recent action probability ratios from the importance weight of the per-decision importance sampling technique, where the number $k$ is selected based on a covariance test. Removing time steps in this manner is not always a natural solution since the $i$'th decision in one trajectory may not be related to the $i$'th decision in another trajectory. In state-based importance sampling, one drops action probability ratios associated with a particular set of states rather than a particular set of time steps. In doing so, a similar covariance test is proposed as one technique to identify the set of states to drop. 

It is also possible to avoid products of action probability ratios in favour of sums over state density ratios. Stationary density ratio estimation (SDRE) \cite{Liu2018} avoids the cumulative product across time and instead is based on the average visitation distribution of state-action pairs. Denoting the average visitation distribution of a state $s_t^{(i)}$ as $d_{t}^{\pi}(s_t^{(i)})$ and its estimate as  $\hat{d}_t^{\pi_e}(s_t^{(i)})$, SDRE estimates the return as
\begin{equation}
\label{eq: SDRE}
\hat{G}_{SDRE} = \frac{1}{n} \sum_{i=1}^{n}  \sum_{t=1}^{H} r_t^{(i)} \frac{\hat{d}^{\pi_e}(s_t^{(i)})}{\hat{d}^{\pi_b}(s_t^{(i)})}  \,,
\end{equation}
thereby removing the curse of horizon such that the variance depends polynomially, rather than exponentially, on $H$. The approach is thereby suitable for large and even infinite-horizon problems but requires additional assumptions of ergodicity, a finite state space, and a clearly defined stationary distribution. In similar spirit, Marginalised importance sampling (MIS) \cite{Xie2019} is formulated as
\begin{equation}
\hat{G}_{MIS} = \frac{1}{n} \sum_{i=1}^{n}  \sum_{t=1}^{H} \hat{r}_t^{\pi_e}(s_t^{(i)}) \frac{\hat{d}_t^{\pi_e}(s_t^{(i)})}{\hat{d}_t^{\pi_b}(s_t^{(i)})} \,,
\end{equation}
where one can note the difference with SDRE by indexing the state distribution $\hat{d}_t^{\pi_e}$ at a particular time $t$ as well as estimating the reward  $\hat{r}_t^{\pi_e}(s_t^{(i)})$ for a particular time $t$. The estimator of the state distribution is updated according to $\hat{d}_t^{\pi_e} = \hat{\mathcal{T}}_{t}^{\pi_e}(s_t \vert s_{t-1}) \hat{d}_{t-1}^{\pi_e}$, where $\hat{\mathcal{T}}_{t}^{\pi_e}(s_t \vert s_{t-1})$ is the empirical importance-weighted visitation of $s_{t-1} \to s_t$. The estimator $\hat{r}_t^{\pi_e}$ is the importance-weighted reward for transitions $s_{t-1} \to s_t$. Within the formalism of a tabular MDP, i.e. where state space and actions space are limited such that each state-action pair can be frequently visited, Tabular MIS \cite{Yin2020} provides an alternative approximation to $\hat{r}_t^{\pi_e}(s_t^{(i)})$ and $\hat{d}_t^{\pi_e}(s_t^{(i)})$ based on the observed visitation frequencies. Tabular MIS provides a lower MSE bound in line with the Cramer-Rao bound of $\Omega(H^2/n)$.  Compared to techniques using state density ratios, which require stationary state distributions and various function approximators, state-based estimators are applicable under varying asssumptions and algorithmic frameworks, depending on which estimator it is plugged into (e.g. ordinary importance sampling vs stationary density ratio estimation).

Conditional importance sampling (CIS) uses conditional Monte Carlo over the importance weight to yield lowered variance. Denoting $\rho_{1:t'} = \prod_{t=1}^{t'} \frac{\pi_e(a_t|s_t)}{\pi_b(a_t|s_t)}$ and $\phi_t$ as a statistic obtained from the trajectory, CIS can be written as
\begin{equation}
\hat{G}_{CIS} = \sum_{i=1}^{n}\sum_{t=1}^{H} r_{t}^{(i)} \mathbb{E}_{\pi_b}[\rho_{1:H}\vert \phi_t]] \,,
\end{equation} 
which yields $\text{Var}(G\mathbb{E}_{\pi_b}[\rho_{1:H}\vert \phi_t]) \leq \text{Var}(G\rho_{1:H})$. Liu et al. (2020)
\cite{Liu2020a} make use of this property to better understand IS, PDIS, and marginalised IS, noting that in this framework the PDIS estimate has $\phi_t = s_{1:t},a_{1:t}$ and the marginalised IS estimate has $\phi_t = s_t, a_t$. Rowland et al. (2020) \cite{Rowland2019} present a related framework for conditional importance sampling with emphasis on trajectories given particular start state-action pair. They introduce novel algorithms arising within the framework, namely return-conditioned, reward-conditioned, and state-conditioned importance sampling. When the equations cannot be solved analytically, the technique often comes with a requirement to form an importance weight regression model, which may increase variance and reduce accuracy. Rather than conditioning on states, state-based methods drop selected states from the importance weight every time they occur in the trajectory.

State-based techniques can also be applied when an accurate model is available. In particular, doubly robust (DR) methods \cite{Dudik2011,Jiang2016,Farajtabar2018,Thomas2016} use estimators that combine importance sampling with the direct method, which uses an estimated model to infer the return of the evaluation policy, using
traditional statistical techniques for missing data (e.g. \cite{Bang2005}). The doubly robust estimator benefits from the low variance of the direct method and the low bias -- or unbiasedness, if the behaviour policy is known -- of importance sampling methods. It is a consistent estimator \cite{Thomas2016}, implying its variance converges to 0 for $n \to \infty$.

Other variance reduction techniques not investigated in this paper are of further interest, including options \cite{Sutton1999,Guo2017}, multiple importance sampling \cite{Owen2013,Metelli2020}, and estimator selection \cite{Udagawa2022}. 

\subsection{Applying off-policy evaluation}
Off-policy evaluation has two primary uses. First, when few trajectories are available and the designer wants to make a decision on whether or not to 
use the evaluation policy instead of a given expert policy; primary examples include medical treatment \cite{Chandak2020a} and digital marketing 
\cite{Theocharous2015}, which would incur significant costs and safety risks when the evaluation policy's performance is not estimated correctly. 
Second, off-policy evaluation can be used to optimise the policy, where key challenges include how to provide monotonic improvement and how to provide confidence estimates on improvement. For example, high-confidence policy improvement (HCOPE) \cite{Thomas2015b} uses confidence 
estimates on the evaluation \cite{Thomas2015a} to ensure with high probability that the new policy improves on the old; policy optimisation via 
importance sampling (POIS) \cite{Papini2019} optimises surrogate objective that effectively captures the trade-off between the estimated improvement and the uncertainty due to importance sampling \cite{Metelli2020}; and Uniform OPE \cite{Yin2021} simulateneously evaluates all policies in a policy class.

\section{State-based Importance Sampling}
\label{sec: SIS}
State-based importance sampling (SIS) mitigates the above issue by constructing an estimator $\hat{G}_{SIS}$ that selectively drops
the action probability ratios of a select state set $\mathcal{S}^A \subset \mathcal{S}$ from the product to compute the importance weight. If the dropped term is equal to 1 and does not covary with the estimator $\hat{G}_{SIS}$, this yields lower-variance estimates. In this section, we demonstrate the state-based formalism for ordinary importance sampling while Sec.~\ref{sec: variants} shows how to apply the state-based formalism for other estimators with lower variance and mean squared error. We extend ordinary importance sampling here for its simplicity and the fact that its variance is more significantly reduced (i.e. by reducing the exponent).

Let $\mathcal{S}^A \subset \mathcal{S}$ and $\mathcal{S}^B = \mathcal{S} \setminus \mathcal{S}^A$ its complement. Let $G = \sum_{t=1}^{H} r_t$. Also for any given trajectory $\tau = \{s_1,a_1,\dots,s_H,a_H\}$ taken by a policy $\pi_b$ on an MDP $\mathcal{M}$, let 
\begin{align*}
A = \begin{cases}
\prod_{t\in [H]: s_t \in \mathcal{S}^A} \frac{\pi_e(a_t|s_t)}{\pi_b(a_t|s_t)} , & \text{if $\exists s \in \mathcal{S}^A \cap \tau$}\\
1,& \text{otherwise} \\
\end{cases} \\
B = 
 \begin{cases} \prod_{t\in [H]: s_t \in \mathcal{S}^B} \frac{\pi_e(a_t|s_t)}{\pi_b(a_t|s_t)} , & \text{if $\exists s \in \mathcal{S}^B \cap \tau$}\\
 1,& \text{otherwise,} \\
 \end{cases}
\end{align*}
where $[H] = \{1,2,\dots,H\}$.   

Given a selected state-set $\mathcal{S}^A$ of dropped states, the state-based variant of ordinary importance sampling computes the action probability ratios only over its complement $\mathcal{S}^B$:
\begin{equation}
\hat{G}_{SIS}(\mathcal{S}^A) = \frac{1}{n} \sum_{i=1}^{n}  G^{(i)} \left( \prod_{t\in [H]: s_t^{(i)} \in \mathcal{S}^B} \frac{\pi_e(a_t^{(i)}|s_t^{(i)})}{\pi_b(a_t^{(i)}|s_t^{(i)})} \right) \,.
\end{equation}
Its expected return can be rewritten as
\begin{align}
\label{eq: decomposition}
\mathcal{G} &= \mathbb{E}_{\pi_b}\left[ A \right] \mathbb{E}_{\pi_b}\left[ BG \right]  + \text{Cov}(A,BG) \,, 
\end{align}
since $\text{Cov}(X,Y) = \mathbb{E}[XY] - \mathbb{E}[X]\mathbb{E}[Y]$. As shown in the coviarance testing framework \cite{Guo2017}, any decomposition of the form \eqref{eq: decomposition} allows to accurately estimate $\mathcal{G}$ based on samples of $BG$  if $\mathbb{E}_{\pi_b}\left[ A \right] \approx 1$ and $\text{Cov}(A,BG)$ is sufficiently small. For the first condition, the random variable $A$ can be seen as the importance weight under a different evaluation policy $\pi_e^A$, which we will call the \textbf{$\mathcal{S}^A$-reduced evaluation policy} of $\pi_e$, which is equal to $\pi_e$ for $s \in \mathcal{S}^A$ and equal to $\pi_b$ for $s \in \mathcal{S}^B$;  therefore it follows that $\mathbb{E}[A] = 1$ (see Appendix~A). The second condition does not follow automatically, but depends on the chosen $\mathcal{S}^A$; the definition below specifies the criterion for $S^A$ to form an accurate state-based estimator.
\begin{definition}
\label{def: negligible}
\textbf{$(\pi_e,\pi_b,\epsilon)$-decomposable MDP and $\epsilon$-negligible state set.} Let $\epsilon > 0$, $\mathcal{M}$ be an MDP $\langle \mathcal{S},\mathcal{A},  r, \mathcal{T} \rangle$, and let $\pi_e$ and $\pi_b$ two policies in $M$. Then $\mathcal{M}$ is $(\pi_e,\pi_b,\epsilon)$-decomposable iff there exists a set $\mathcal{S}^A$ for which $\text{Cov}(A,BG) < \epsilon$.  Any such set $\mathcal{S}^A$ is called an $\epsilon$-negligible state set for the off-policy evaluation problem $\langle \mathcal{M}, \pi_e,\pi_b \rangle$.
\end{definition} 

Applying bias-variance decomposition, it follows that
\begin{equation}
\label{eq: MSE}
\text{MSE}(\hat{G}_{SIS}(\mathcal{S}^A)) = \text{Cov}(A,BG)^2 + \text{Var}(\hat{G}_{SIS}(\mathcal{S}^A)) \,.
\end{equation}
Denoting the subtrajectory $\tau_B  := \{ t\in [H]: s_t^{(i)} \in \mathcal{S}^B \}$ and $M_B := \max_{\tau \in \tau_B} \vert \tau \vert$, we have
\begin{equation}
\label{eq: var}
\text{Var}(\hat{G}_{SIS}(\mathcal{S}^A)) \leq \frac{1}{4n} \left(H r_{\text{max}} \rho_{\text{max}}^{M_B} \right)^2  \,.
\end{equation}
The resulting variance upper bound of $\hat{G}_{SIS}$ is exponential in the maximal number of occurrences of states in $\mathcal{S}^B$ rather than in the full trajectory length $H$. Consequently, the SIS estimator will significantly improve on IS if many visited states can be dropped from the computation of the importance weight. When $\mathcal{S}^A$ is an $\epsilon$-negligible state set, this implies via \eqref{eq: MSE} and \eqref{eq: var} that
\begin{equation}
\text{MSE}(\hat{G}_{SIS}(\mathcal{S}^A)) \leq \epsilon^2 + \frac{1}{4n}\left(H r_{\text{max}} \rho_{\text{max}}^{M_B} \right)^2 \,.
\end{equation}

\section{Optimising the dropped state set}
\label{sec: search}
We propose two search algorithms to optimise the dropped state set, one based on covariance testing and one based on state-action values (a.k.a. Q-values). Additional derivations are presented in Appendix~B showing that even with some additional bias, the performance of a state-based variant of ordinary importance sampling (SIS) is still a significant improvement over ordinary importance sampling.
\subsection{Covariance testing} 
Following the covariance testing approach \cite{Guo2017}, this algorithm applies a bias-variance decomposition based on empirical estimates of the covariance and variance, and compares the MSEs of different parameter choices. In our case, the parameter choice is the state set $S^A$ for a decomposition
\begin{equation}
\widehat{\text{MSE}}(\hat{G}_{SIS}(\mathcal{S}^A)) = \widehat{\text{Cov}}(A,BG)^2 + \widehat{\text{Var}}(\hat{G}_{SIS}(\mathcal{S}^A)) \,,
\end{equation}
and the algorithm encourages larger negligible states sets using the criterion $\texttt{best} \gets \mathcal{S}^A$ if and only if $\widehat{\text{Cov}}(A,BG) < \epsilon$ and $\widehat{\text{MSE}}(\mathcal{S}^A) < \widehat{\text{MSE}}(\texttt{best})$ or $\widehat{\text{MSE}}(\mathcal{S}^A) < \widehat{\text{MSE}}(\texttt{best})*(1+\epsilon_s) \land  \vert \mathcal{S}^A  \vert > \vert \texttt{best}  \vert$ based on  user-defined parameters $\epsilon_s>0$ and $\epsilon>0$. In this paper, we use exhaustive search over subsets while it is also possible to form heuristic search algorithms.

\subsection{Q-value based identification} 
In the Q-value based identification, a model of the MDP is formulated based on the trajectories from $\pi_b$ to compute time-specific Q-values for $\pi_e$ via stochastic dynamic programming:
\begin{equation}
\label{eq: Q}
\hat{Q}_t(s,a) = \hat{r}(s,a) + \sum_{s' \in \mathcal{S}} \hat{\mathcal{T}}(s,a,s') \pi_e(: \vert s') \cdot \hat{Q}_{t+1}(s',:) \,,
\end{equation}
where $:$ indicates vectorisation. The algorithm identifies a state $s \in \mathcal{S}$ as negligible if for all $t \in [H]$,  $\vert \hat{Q}_t(s,a) - \hat{Q}_t(s,a')\vert  < \epsilon$ for all $a,a' \in \mathcal{A}$. This criterion leads to an SIS estimator with limited absolute bias depending on the user-defined $\epsilon$, the maximal number of dropped weights, and the model approximation error $\epsilon_2$ (see Appendix~C). Consistent with the analysis in Appendix~B, the MSE of SIS is $\mathcal{O}\left(\exp(M_B)\right)$ with both state identification methods. The Q-value based identification algorithm is computationally inexpensive and can include a large subset of the state space as negligible states. Moreover, it easy to interpret and independent of the estimator making it applicable to a variety of estimators.

\begin{table*}[htbp]
\centering
\caption{Mean squared error (MSE) for off-policy evaluation estimators of the expected return in the deterministic lift domain based on 50 independent runs of 1,000 episodes for each domain size. The best and second best estimators are highlighted in bold underlined and bold, respectively, for each domain size. All SIS estimators shown use the Q-value based algorithm for negligible state set identification.}\label{tab: MSE-determ}%
\begin{tabular}{p{1.2cm} | p{1.2cm} p{1.2cm} p{1.2cm} p{1.2cm} p{1.2cm}  p{1.2cm} p{1.2cm}  p{1.2cm}}
\toprule
 \textbf{Domain size} & IS & SIS & PDIS & SPDIS & INCRIS & SINCRIS & SDRE & SSDRE \\ \midrule
7 & \textbf{0.0067} & \underline{\textbf{0.0019}} & 0.0398 & 0.0121 & 4.0024 & 4.0007 & 532.7567 & 536.3787 \\
9 & \textbf{0.0153} & \underline{\textbf{0.0019}} & 0.1766 & 0.0215 & 6.1964 & 6.1765 & 577.3089 & 584.4657 \\
11 & \textbf{0.0353} & \underline{\textbf{0.0029}} & 0.7087 & 0.0600 & 8.9408 & 8.9225 & 674.0772 & 680.1959 \\
13 & 0.0691 & \underline{\textbf{0.0026}} & 1.8038 & \textbf{0.0688} & 12.5475 & 12.4779 & 652.6004 & 660.2387 \\
15 & 0.1389 & \underline{\textbf{0.0033}} & 5.3474 & \textbf{0.1378} & 16.1902 & 16.1558 & 803.3450 & 813.4997 \\
17 & 0.3026 & \underline{\textbf{0.0031}} & 15.1566 & \textbf{0.1303} & 23.7084 & 20.1534 & 758.2237 & 764.6283 \\
\bottomrule
\end{tabular}
\end{table*}

\begin{table*}[htbp]
\centering
\caption{Mean squared error (MSE) for off-policy evaluation estimators of the expected return in the stochastic lift domain based on 200 independent runs of 1,000 episodes for each domain size. The best, second best, and third best estimators are highlighted in bold double-underlined, bold underlined, and bold, respectively, for each domain size.}
\label{tab: MSE-stoch}%
\begin{tabular}{p{1.2cm} | p{1.2cm} p{1.2cm} p{1.2cm} p{1.2cm} p{1.2cm}  p{1.2cm} p{1.2cm}  p{1.2cm} p{1.2cm}  p{1.2cm}}
\toprule
 \textbf{Domain size} & WIS & WSIS & WPDIS & WSPDIS & WINCRIS & WSINCRIS & SDRE & SSDRE & WDR & WDRSIS \\ \midrule
7 & 10.2238 & 8.4832 & 6.8271 & 5.3853 & \underline{\textbf{1.0958}} & \underline{\underline{\textbf{0.2618}}} & 208.7413 & 209.8354 & 5.2572 & \textbf{5.0844} \\
9 & 9.4737 & 7.7326 & 5.3924 & 4.0441 & \underline{\underline{\textbf{0.2973}}} & \underline{\textbf{0.6929}} & 243.1366 & 245.1004 & 2.8067 & \textbf{2.7486}  \\
11 & 9.7285 & 6.2044 & 5.7219 & 3.0552 & \underline{\underline{\textbf{0.2002}}} & 2.3050 & 250.3149 & 251.5843 & \textbf{2.3016} & \underline{\textbf{1.9838}}  \\
13 & 8.4436 & 9.4967 & 4.4449 & 3.2916 & \underline{\underline{\textbf{0.8958}}} & 4.8960 & 266.7911 & 268.6634 & \textbf{1.3459} & \underline{\textbf{1.3354}}  \\
15 & 18.0453 & 4.2831 & 6.2259 & \textbf{1.7004} & 2.6550 & 9.0162 & 280.6434 & 282.6849 & \underline{\textbf{1.2258}} & \underline{\underline{\textbf{0.8797}}}  \\
17 & 8.2474 & 4.8129 & 5.2761 & \textbf{1.8151} & 4.7157 & 13.0056 & 305.7465 & 307.9191 & \underline{\textbf{0.8385}} & \underline{\underline{\textbf{0.4820}}} \\
\bottomrule 
\end{tabular}
\end{table*}

\section{State-based importance sampling variants}
\label{sec: variants}
While so far we have focused on ordinary importance sampling, it is easy to extend SIS to use more advanced off-policy evaluation estimators as its base-estimator.

First, methods such as PDIS and INCRIS, and even the weighted counterparts of these algorithms, are easily extended by analogously setting the action probability ratio to one for negligible states. 

Second, for doubly robust off-policy evaluation \cite{Jiang2016,Thomas2016}, one can use a model using the same formalism as the Q-value based identification. While traditional DR uses an importance weight $w_t^i = \frac{1}{n} \prod_{j=1}^{t} \frac{\pi_{e}(a_t^i \vert s_t^i) }{ \pi_{b}(a_t^i \vert s_t^i)}$ for a trajectory $i$ at time $t$ and $n$ total trajectories, doubly robust state-based importance sampling (DRSIS) uses the importance weight $w_t^i = \frac{1}{n} \prod_{j=1}^{t} \frac{\pi_{e}^A(a_t^i \vert s_t^i)}{\pi_{b}(a_t^i \vert s_t^i)}$ where $\pi_{e}^A$ is the $\mathcal{S}^A$-reduced evaluation policy of $\pi_e$.

Third, techniques that rely on importance sampling over state visitation frequencies such as SDRE and MIS can also benefit from a state-based formulation. For example, the stationary density ratio estimator \eqref{eq: SDRE} can be turned into a state-based formulation according to
\begin{align}
\hat{G}_{SSDRE} = \frac{1}{n} \sum_{i=1}^{n}  \sum_{t=1}^{H} r_t^{(i)} \frac{\hat{d}^{\pi_e^A}(s_t^{(i)})}{\hat{d}^{\pi_b}(s_t^{(i)})}  \,,
\end{align}
where $\hat{d}^{\pi_e^A}(s_t^{(i)})$ is equal to $\hat{d}^{\pi_e}(s_t^{(i)})$ for $s_t^{(i)} \in \mathcal{S}^B$ and $\hat{d}^{\pi_b}(s_t^{(i)})$ for $s_t^{(i)} \in \mathcal{S}^A$. Such techniques are polynomial in $H$ due to replacing the product over action probability ratios by a sum of state visitation density ratios. As such the variance reduction due to reducing polynomial factors is smaller than variance reductions due to reducing the exponent, as is the case in traditional IS methods. However, as the experiments in Sec.~\ref{sec: experiments} show, this can still yield significant MSE improvements and an excellent overall performance in environments with stationary distributions.
\section{Experiments}
\label{sec: experiments}
The experimental validation of state-based estimators comprises 4 domains. The first two domains include lift states in the form of a lift transporting the agent from one side to another, while the third and fourth domain are common RL benchmarks, namely the inventory management domain \cite{Puterman2005} and the taxi domain variant used in \cite{Liu2018}. In each of the domains, we evaluate the expected return of a near-optimal policy $\pi_e$ based on sub-optimal policy $\pi_b$.

The experiments validate state-based importance sampling when implemented on IS, PDIS, INCRIS, DR, and SDRE. The corresponding state-based implementations are denoted with the prefix \textbf{S}; for instance, SPDIS denotes state-based PDIS. State-based doubly robust off-policy evaluation is denoted as \textbf{DRSIS} to distinguish it from SDRE. With the exception of the first domain, which has short trajectory lengths, and SDRE for which this is not applicable, the experiments use additional variance reduction with weighted importance sampling, which is indicated by the prefix \textbf{W}; for instance, \textbf{WIS} denotes weighted IS. Further, the effectiveness of the negligible state identification is demonstrated on the first two domains as these contain known lift states (see Appendix~D). Due to its ability to identify the lift states and yield lower MSE, the Q-based identification algorithm is used throughout this section. Additional experimental details are provided in Appendix~E and source code  is available at \url{https://github.com/bossdm/ImportanceSampling}.

\begin{table*}[htbp]
\centering
\caption{Normalised mean squared error (MSE) for off-policy evaluation estimators of the expected return in the inventory management domain based on 50 independent runs for each number of episodes. Normalisation is based on division by the squared range. The best, second best, and third best estimators are highlighted in bold double-underlined, bold underlined, and bold, respectively, for each number of episodes.}\label{tab: MSE-IM}
\begin{tabular}{p{1.2cm} | p{1.2cm} p{1.2cm} p{1.2cm} p{1.2cm} p{1.2cm} p{1.2cm}  p{1.2cm} p{1.2cm} p{1.2cm} p{1.2cm}}
\toprule
 \textbf{Episodes} & WIS & WSIS & WPDIS & WSPDIS & WINCRIS & WSINCRIS & SDRE & SSDRE & WDR & WDRSIS \\ \midrule
100& 0.0490 & \textbf{0.0088} & 0.0278 & \underline{\underline{\textbf{0.0074}}} & 0.0641 & 0.0109 & 0.0160 & \underline{\textbf{0.0077}} & 0.0173 & 0.0138 \\
250& 0.0651 & \textbf{0.0080} & 0.0335 & \underline{\textbf{0.0080}} & 0.0546 & 0.0118 & 0.0161 & \underline{\underline{\textbf{0.0070}}} & 0.0132 & 0.0137 \\
500& 0.0652 & 0.0113 & 0.0340 & \textbf{0.0105} & 0.0456 & 0.0124 & 0.0163 & \underline{\underline{\textbf{0.0067}}} & \underline{\textbf{0.0098}} & 0.0114 \\
1000& 0.0820 & 0.0122 & 0.0293 & 0.0119 & 0.0313 & 0.0128 & 0.0162 & \underline{\underline{\textbf{0.0066}}} & \textbf{0.0117} & \underline{\textbf{0.0104}} \\
\bottomrule
\end{tabular}
\end{table*} 
\begin{table*}[htbp]
\centering
\caption{Mean squared error (MSE) for off-policy estimators of the expected return in the taxi domain based on 20 independent runs for each effective horizon. The best, second best, and third best estimators are highlighted in bold double-underlined, bold underlined, and bold, respectively, for each effective horizon.}\label{tab: MSE-taxi}
\begin{tabular}{p{1.2cm} | p{1.2cm} p{1.2cm} p{1.2cm} p{1.2cm} p{1.2cm} p{1.2cm}  p{1.2cm} p{1.2cm} p{1.2cm} p{1.2cm}}
\toprule
\textbf{Effective horizon} & WIS & WSIS & WPDIS & WSPDIS & WINCRIS & WSINCRIS & SDRE & SSDRE & WDR & WDRSIS \\ \midrule
10 & 0.6080 & \underline{\underline{\textbf{0.4525}}} & 0.7234 & 0.5084 & \textbf{0.4896} & \underline{\textbf{0.4767}} & 1.2764 & 1.2790 & 53.2003 & 53.1011 \\
50 & 15.8227 & 15.9281 & 11.6239 & \textbf{10.7777} & 35.6350 & 35.8282 & \underline{\textbf{9.4938}} & \underline{\underline{\textbf{9.4435}}} & 188.8620 & 187.3876 \\
250 & 440.6435 & 473.4986 & 281.9518 & \textbf{273.3647} & 1753.7938 & 1752.6067 & \underline{\textbf{25.3142}} & \underline{\underline{\textbf{25.2920}}} & 993.3100 & 1055.8603 \\
1000 & 8233.7773 & 7948.8306 & 4876.1759 & \textbf{4652.7797} & 27675.6630 & 27635.4627 & \underline{\textbf{231.6101}} & \underline{\underline{\textbf{231.4313}}} & 11982.1668 & 12176.5372 \\
\end{tabular}
\end{table*}
\subsection{Lift domains}
Based on the $\epsilon$-negligibility definition, one would expect state-based techniques to be advantageous on domains where there are some states for which the action taken has no or limited impact on the return. This intuition is captured by the notion of ``lift states''. 
\begin{definition}
\textbf{Lift states.} A lift state $s \in \mathcal{S}$ is a state for which $\mathcal{T}(s,a) = \mathcal{T}(s,a')$  and $r(s,a) = r(s,a')$ for all $a,a' \in \mathcal{A}$. The set of lift states for a domain is denoted by $\mathcal{S}^L$.
\end{definition}

\paragraph{Deterministic lift domain}
The deterministic lift domain involves an agent moving on a straight line, with the state space $\mathcal{S} = \{-b,\dots,-1,0,1,\dots,b\}$ representing the coordinates along the line and the action space $\mathcal{A}=\{\texttt{left},\texttt{right} \}$ representing left or 
right movements on the line. With the exception of the starting state and states within a distance of 1 from the terminal state, the domain has lift states which force the agent to move left on $s < 0$ and right on $s > 0$. The episode stops when the agent hits the 
bound, and then the terminal reward of $r=s$ is given (equal to either $-b$ or $b$). All non-terminal rewards are $r=-1$ to penalise the length of the path. The experiments 
manipulate this bound for an increasing effective horizon. An analysis comparing state-based estimators  to their traditional counterparts (see Tab.~\ref{tab: MSE-determ}) demonstrates a 
near universal benefit in the deterministic domain, with the only exception being SDRE. SIS is the best performing estimator across all domain sizes, followed by IS and SPDIS depending on the 
domain size. SDRE and SSDRE have particularly poor estimates because the domain does not have a stationary distribution. Analysis of the residuals (see Fig.~1a in Appendix~F) also shows that: a) state-based methods have successfully reduced the variance compared to their traditional counterparts; b) SPDIS 
 strongly reduces the residual of PDIS for large domain sizes; and c) INCRIS and SINCRIS have large residuals, which implies that dropping recent time steps may not be suitable for the deterministic domain (e.g. the first time step always determines the final reward). Due to the deterministic domain being easy to model, the doubly robust methods have MSEs close to zero in this domain so are not further analysed.

\paragraph{Stochastic lift domain}
In the stochastic lift domain, the evaluation policy and the state transitions are 
stochastic rather than deterministic. The evaluation policy $\pi_e$ takes the best action with probability $1-\delta$ and the other action with 
probability $\delta$, and the state transitions dynamics are formulated such that the action effect is reversed with probability $\delta$. An analysis comparing state-based estimators  to their traditional counterparts (see Tab.~\ref{tab: MSE-stoch}) demonstrates a benefit in the stochastic domain. This time doubly robust methods are also analysed as the error of the DR estimator is now non-zero. WSIS and WSPDIS consistently outperform their traditional counterparts. For large domain sizes, WDRSIS is the best performing estimator followed by WDR and WSPDIS. State-based methods are not adversely affected by the domain size, which is explained by all the lift states being dropped and therefore the maximal number of steps $M_B$ in non-negligible states remaining constant regardless of domain size. For small domain sizes, WINCRIS and WSINCRIS have the best performance, indicating the benefit of dropping time steps; the difference to the deterministic domain maybe due to the visitation of multiple lifts in the same episode. The lower performance of WINCRIS and WSINCRIS in larger domain sizes is attributed to the limited history window  $k \in \{1,\dots,10\}$, which is needed for computational feasibility but may not be sufficient for longer trajectories. Fig.~1b in Appendix~F further demonstrates that the state-based methods have successfully reduced the variance compared to their traditional counterparts. As in the deterministic lift domain, the poor performance of SDRE estimators in the stochastic lift domain is attributed to the domain not having a defined stationary state distribution.

\subsection{Inventory management}
The task of the agent in the inventory management problem is 
to purchase items to make optimal profits selling the items. The agent must balance supply with demand while having at most $S$ items in the inventory. The state is the current inventory while the action is the chosen number of items to add. States corresponding to an almost full inventory  are similar to lift states in the sense that the actions chosen are unimportant. As the domain is high variance due to the varying demand, we opt for a relatively high number of 50 experiment runs and instead of manipulating the effective horizon we analyse the improvement in performance across different numbers of episodes. An analysis comparing state-based estimators  to their traditional counterparts (see Tab.~\ref{tab: MSE-IM}) demonstrates a near universal benefit in the inventory management domain, with 2-- to 8--fold improvements being observed (see Tab.~\ref{tab: MSE-IM}) -- with the exception of DR which already has high accuracy. WSPDIS performs best for 100 episodes while SSDRE performs best for 250 to 1,000 episodes. The high performance of SSDRE is attributed to a clearly defined stationary distribution. The performance of SSDRE and SDRSIS dependent on the sample size suggests that these state-based methods provide improved convergence and efficiency. Other high-performing algorithms are WSIS, WSINCRIS, WDR, and WDRSIS. Fig.~2 in Appendix~F demonstrates that state-based techniques reduce the variance and yield a smaller residual.

\subsection{Taxi}
The taxi domain has a much larger state space with $\vert \mathcal{S} \vert = 2000$ and can have infinite horizon. A taxi agent has to pick up and drop off passengers who arrive at random corners in a $5 \times 5$ grid based on $6$ actions, namely going a step north, west, south, east, picking up a passenger, and dropping off a passenger. The taxi agent receives a reward of $20$ upon a drop-off or pick-up at the correct location, and a reward of $-1$ otherwise. The domain is of interest to investigate the scalability to large state-action spaces and large effective horizon, and it is of interest to manipulate the effective horizon in an additional domain which in contrast to the lift domains has a stationary distribution. The results for the taxi domain (see Tab.~\ref{tab: MSE-taxi}) also show a near universal benefit for state-based estimators compared to the traditional counterparts, although the effect is not as pronounced as for example in the inventory management domain.  
For an effective horizon of 10, WSIS is the highest performing followed by WSINCRIS and WINCRIS. For effective horizons between 50 and 1,000,  SSDRE is the highest performer followed by SDRE and WSPDIS. The high performance of SDRE based methods, and their relative improvement across effective horizon, is attributed to their MSE being polynomial in the effective horizon whereas the MSE of traditional importance sampling techniques based on products of action probability ratios is exponential in the effective horizon. As illustrated in Fig.~3 in Appendix~F, the SDRE methods have a lower variance than WDR methods and their residuals do not depend as strongly on the effective horizon as WIS, WPDIS, and WINCRIS.

\section{Conclusion}
Off-policy reinforcement learning comes with the significant challenge of high-variance estimates for the expected return. This paper introduces 
state-based importance sampling, a class of off-policy evaluation techniques which reduces the variance of importance sampling by eliminating states that do not affect the return from the importance weight computation. The state-based technique is shown to improve performance when integrated with a variety of off-policy evaluation methods,  making it a useful plug-in for many applications. Future work directions include theoretical studies on the presence of negligible states, implementing policy optimisation techniques based on state-based importance sampling, and exploring the links with other off-policy evaluation techniques.

\section*{Acknowledgement}
We thank Shreyas Chaudhari for comments on a draft of the paper.
\bibliographystyle{ieeetr}
\bibliography{biblio}

\clearpage
\onecolumn
{\huge{Supplementary Information}}

\section*{Appendix A: Proof of $\mathbb{E}[A]=1$}
Denoting $P_1$ as the starting distribution, $\tau := s_1,a_1\dots, s_H,a_H$, $\tau_s := s_1, \dots, s_H$, and $\tau_a := a_1, \dots, a_H$ and $\pi_e^A$ as the $\mathcal{S}^A$-reduced evaluation policy of $\pi_e$, we have
\begin{align*}
\mathbb{E}_{\pi_b}\left[ A \right] &= \int_{\tau} \pi_e^A(\tau_a \vert \tau_s)/\pi_b(\tau_a\vert \tau_s) \mathbb{P}(\tau_s \vert \pi_b) d\tau \\
							       &= \int_{\tau} \pi_e^A(\tau_a\vert \tau_s)/\pi_b(\tau_a\vert \tau_s) \\ & \hspace{1cm} P_1(s_1)  \prod_{t=1}^{H} \pi_b(a_t \vert s_t) \mathcal{T}(s_t,a_t,s_{t+1}) \\
							        &= \int_{\tau} \pi_e^A(\tau_a\vert \tau_s) P_1(s_1)  \prod_{t=1}^{H} \mathcal{T}(s_t,a_t,s_{t+1}) d\tau \\
							        &= \int_{\tau} \mathbb{P}(\tau_s \vert \pi_e^A) d\tau  \\
							        &= 1 \,.
\end{align*}

\section*{Appendix B: Analysis of MSE under inaccurate state-set identification}
The variance reduction of state-based methods depends crucially on the state-set identification. In covariance testing, the covariance estimate may be highly inaccurate, and in Q-value based identification, the model on which the Q-value is based may be highly inaccurate. Below we compare the true MSE of the IS estimator to that of the covariance testing based SIS estimator under the assumption that the state-identification algorithm has identified a state-set that is not negligible at all but instead has an arbitrary covariance with the return estimate (thereby yielding a significant bias in the estimate).

Note that the true MSE of ordinary importance sampling is upper bounded by
\begin{align*}
\text{MSE}(\hat{G}_{IS}) &\leq \frac{1}{4n} \left(H r_{\text{max}} \rho_{\text{max}}^H\right)^2 \\
						 &= \mathcal{O}(\exp(H)) \,.
\end{align*}
While for SIS the chosen state-set has been observed to be $\epsilon$-negligible and the covariance and variance have been estimated, for the sake of the argument assume that the population covariance and variance may take on any value. Via Cauchy-Schwartz inequality and Popoviciu's inequality with $G \in [0,Hr_{\text{max}}]$ and $A, B \in [0,\rho_{\text{max}}]$, we have
\begin{align*}
\vert \text{Cov}(A,BG) \vert \leq \text{Var}(A)\text{Var}(BG) 
							\leq \frac{\rho_{\text{max}}^4 H^2 r_{\text{max}}^2}{16} \,.
\end{align*} 
With the bias equal to the covariance, it follows via bias-variance decomposition that
\begin{align*}
\text{MSE}(\hat{G}_{SIS}(\mathcal{S}^A)) &= \text{Var}(\hat{G}_{SIS}(\mathcal{S}^A)) + \text{Cov}(A,BG)^2   \\
										 &\leq \frac{1}{4n} \left(H r_{\text{max}} \rho_{\text{max}}^{M_B} \right)^2  + \frac{\rho_{\text{max}}^8 H^4 r_{\text{max}}^4}{256} \\
										 &= \mathcal{O}(\exp(M_B)) \,.				
\end{align*}
Therefore the SIS has a significantly improved upper bound on the true MSE provided the maximal number of steps in $S^B$ is significantly smaller than $H$. If, however, the state-set identification was accurate,  such that $\text{Cov}(A,BG) < \epsilon$, then any reduction in variance implies via the bias-variance decomposition that SIS has successfully optimised the true MSE compared to IS. 

\section*{Appendix C: Proof of limited bias of Q-based state identification}
Let $s_{t} \in S^A = \{ s \in \mathcal{S}: \vert \hat{Q}_{t'}(s,a) - \hat{Q}_{t'}(s,a')\vert  < \epsilon \quad \forall a,a' \in \mathcal{A} \quad \forall t' \in [H] \}$. This implies that the SIS estimator has the following expected return from starting state $s_t$
\begin{align*}
\mathbb{E}_{\pi_b}[\hat{G}_{SIS} \vert s_t] &= \mathbb{E}_{\pi_b}[\rho_{t+1:t+H} G_t \vert s_t] \\
 &= \sum_{a} \pi_b(a \vert s_t) \left( r(s_t,a) + \mathbb{E}_{\pi_b}[\rho_{t+1:t+H} G_{t+1} \vert s_t] \right) \\
    &= \sum_{a} \pi_b(a \vert s_t) Q_{t}(s_t,a)  \\
    &\approx \sum_{a} \pi_e(a \vert s_t) Q_{t}(s_t,a)  \\
&= \mathbb{E}_{\pi_e}[G_t \vert s_t] \,,
\end{align*} 
where the approximate equality follows since $\vert \hat{Q}_t(s,a) - \hat{Q}_t(s,a')\vert  < \epsilon$ implies $Q_t(s,a) \approx Q_t(s,a')$ for all $a,a' \in \mathcal{A}$. Under a perfect model, the bias of the estimator comes only from the approximate equality, in which case the absolute bias induced by state being dropped is at most $\epsilon$. Under an erroneous model, which yields $\vert \hat{Q}_t(s,a) - Q_t(s,a) \vert < \epsilon_2$ for all $t$ and all $a \in \mathcal{A}$, the absolute bias induced by a state being dropped is at most $\epsilon + \epsilon_2$.  Repeatedly applying this argument for a maximal number of $M_A$ dropped weights in a trajectory, the absolute bias induced is at most $M_A(\epsilon + \epsilon_2)$. Filling in $\mathcal{O}\left(\text{poly}(H)\right)$ as the bias and $\frac{1}{4n} \left(H r_{\text{max}} \rho_{\text{max}}^{M_B} \right)^2$ as the variance in the bias-variance decomposition, the MSE is bounded by $\mathcal{O}\left(\exp(M_B)\right)$. \qed

\section*{Appendix D: Comparison of state set identification algorithms}
The comparison for traditional importance sampling includes \textbf{SIS (Lift states)}, which uses the lift states of the domain (i.e., where $\mathcal{S}^A = \mathcal{S}^L$); \textbf{SIS (Covariance testing)}, which uses the negligible states identified by covariance testing; and \textbf{SIS (Q-based)}, which uses the negligible states identified using Q-values. Similarly, we evaluate the effectiveness of state identification for the doubly robust off-policy evaluation method, which are analgously denoted as \textbf{SDR (Lift states)}, \textbf{SDR (Covariance testing)}, and \textbf{SIS (Q-based)}.

Analysing the different negligible state set identification methods reveals that SIS (Q-based) identifies the lift states as negligible states and 
therefore yields the same performance as SIS (Lift states). Tab.~\ref{tab: MSE-SIS-determ} demonstrates the superior performance of SIS 
(Q-based) and SIS (Lift states) compared to IS, which never drops any states, and SIS (Covariance testing), which typically drops two lift states (the 
maximal cardinality given the computational budget).

\begin{table}[htbp!]
\centering
\caption{Mean squared error (MSE) comparing state-based importance sampling methods with different negligible state sets on 50 independent runs of the deterministic lift domain for each domain size. Estimates are obtained based on 1,000 episodes. The best and second best estimators are highlighted in bold underlined and bold, respectively, for each number of episodes; in case of a tie only the best estimators are highlighted in bold underlined.}\label{tab: MSE-SIS-determ}%
\begin{tabular}{l | p{1.6cm} p{1.6cm} p{1.6cm} p{1.6cm}}
\toprule 
& IS & SIS \newline (Lift states) & SIS \newline (Covariance testing) & SIS \newline (Q-based) \\ 
\midrule 
 \textbf{Domain size}& & & & \\
7 & 0.0067 & \underline{\textbf{0.0019}} & 0.0056 & \underline{\textbf{0.0019}} \\
9 & 0.0153 & \underline{\textbf{0.0019}} & 0.0138 & \underline{\textbf{0.0019}} \\
11 & 0.0353 & \underline{\textbf{0.0029}} & 0.0345 & \underline{\textbf{0.0029}} \\
13 & 0.0691 & \underline{\textbf{0.0026}} & 0.0705 & \underline{\textbf{0.0026}} \\
15 & 0.1389 & \underline{\textbf{0.0033}} & 0.1386 & \underline{\textbf{0.0033}} \\
17 & 0.3026 & \underline{\textbf{0.0031}} & 0.3035 & \underline{\textbf{0.0031}} \\
\bottomrule  
\end{tabular}
\end{table}

Results for state set identification are analogous to the deterministic lift domain: WSIS (Q-based) is successful at identifying the lift states as negligible states and together with WSIS (Lift states) has a superior performance compared to WSIS (Covariance testing) and WIS (see Tab.~\ref{tab: MSE-SIS-stoch}; and WSIS (Covariance testing) typically drops two lift states (the maximal cardinality given the computational budget). Tab.~\ref{tab: MSE-DRSIS-stoch} demonstrates how dropping 
states is also beneficial for doubly robust methods and again the Q-based state identification works to identify the lift states. 
\begin{table}[htbp!]
\centering
\caption{Mean squared error (MSE) comparing state-based importance sampling methods with different negligible state sets on 200 independent runs of the stochastic lift domain for each domain size. Estimates are obtained based on 1,000 episodes. The best and second best estimators are highlighted in bold underlined and bold, respectively, for each domain size; in case of a tie only the best estimators are highlighted in bold underlined.}\label{tab: MSE-SIS-stoch}%
\begin{tabular}{l | p{1.6cm} p{1.6cm} p{1.6cm} p{1.6cm}}
\toprule 
 & WIS & WSIS \newline (Lift states) & WSIS \newline (Covariance testing) & WSIS \newline (Q-based) \\ 
 \midrule 
 \textbf{Domain size}& & & & \\
7 & 10.2238 & \underline{\textbf{8.4832}} & 10.2186 & \underline{\textbf{8.4832}}  \\
9 & 9.4737 & \underline{\textbf{7.7326}} & 9.4489 & \underline{\textbf{7.7326}}  \\
11 & 9.7285 & \underline{\textbf{6.2044}} & 9.6775 & \underline{\textbf{6.2044}} \\
13 & \textbf{8.4436} & 9.5003 & \underline{\textbf{8.4058}} & 9.4967 \\
15 & 18.0453 & \underline{\textbf{4.2803}} & 17.9633 & \textbf{4.2831} \\
17 & 8.2474 & \underline{\textbf{4.7482}} & 7.7646 & \textbf{4.8129} \\
\bottomrule  
\end{tabular}
\end{table}

\newpage

\begin{table}[htbp!]
\centering
\caption{Mean squared error (MSE) comparing doubly robust methods with different negligible state sets on 200 independent runs of the stochastic lift domain for each domain size. Estimates are obtained based on 1,000 episodes. The best and second best estimators are highlighted in bold underlined and bold, respectively, for each domain size; in case of a tie only the best estimators are highlighted in bold underlined.}\label{tab: MSE-DRSIS-stoch}%
\begin{tabular}{l | p{1.6cm} p{1.6cm} p{1.6cm} p{1.6cm}}
\toprule
 & WDR & WDRSIS \newline (Lift states) & WDRSIS \newline (Covariance testing) & WDRSIS \newline (Q-based) \\\midrule 
 \textbf{Domain size}& & & & \\
7 & 5.2572 & \underline{\textbf{5.0844}} & 5.2662 & \underline{\textbf{5.0844}}   \\
9 & 2.8067 & \underline{\textbf{2.7486}} & 2.8235 & \underline{\textbf{2.7486}}   \\
11 & 2.3016 & \underline{\textbf{1.9838}} & 2.3058 & \underline{\textbf{1.9838}}   \\
13 & 1.3459 & \textbf{1.3359} & 1.3612 & \underline{\textbf{1.3354}} \\
15 & 1.2258 & \underline{\textbf{0.8788}} & 1.1675 & \textbf{0.8797} \\
17 & 0.8385 & \underline{\textbf{0.4817}} & 0.6620 & \textbf{0.4820} \\
\bottomrule 
\end{tabular}
\end{table}

\section*{Appendix E: Experimental details}
In the experiments, covariance testing is applied with $\epsilon_s = \epsilon = 0.01$ and rather than the full power set, we limit the search to sets with cardinalities of at most 2. The Q-value based identification is suitable for all the domains and the parameter settings are $\epsilon = 1$ in the grid world lift domains, $\epsilon=50$ in the inventory management domain, and $\epsilon=2$ in the taxi domain. 

In the lift domains, $\pi_e$ is the optimal policy and $\pi_b$ is the uniform random policy. The expected return is 1.0 in the deterministic domain and negative in the stochastic domain (with magnitude increasing with domain size). The failure probability is set to $\delta=0.05$ for the stochastic domain.

In inventory management, the purchasing cost is set to $2.49$, the sale price to $3.99$, 
and the holding cost to $0.03$. Demand for items follows a Gaussian with mean $S/4$ and standard deviation $S/6$, and $S=10$ in the experiments. Each episode consists of 100 steps, and consistent with the rest of our experiments an undiscounted setting ($\gamma=1$) is considered. The evaluation policy, $\pi_e$, was trained for 500 episodes on the MDP and has an expected return of $\mathcal{G}=1471.1041$. The behaviour policy, $\pi_b$, is uniform random. 

In taxi, the implementation of the domain is based on code of \cite{Liu2018} (see \url{{https://github.com/zt95/infinite-horizon-off-policy-estimation/tree/master/taxi}}). The evaluation policy represents a policy formed after 1,000 iterations of Q-learning and the behaviour policy represents a policy formed after 950 iterations of Q-learning. The expected return is positive and increases with effective horizon.
\clearpage
\section*{Appendix F: Residual plots}
\begin{figure}[htbp!]
\centering
\subfloat[Deterministic lift domain]{
\includegraphics[width=0.55\linewidth]{{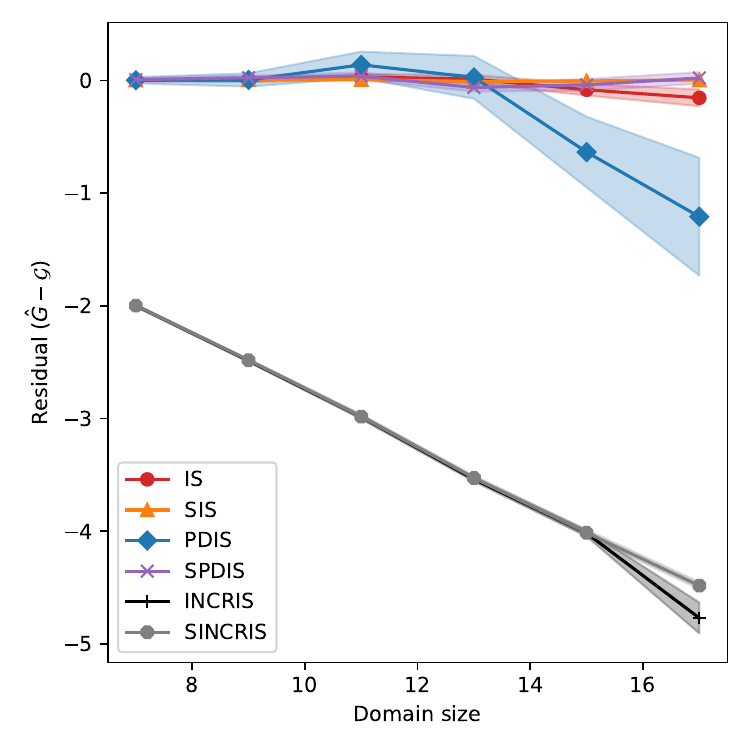}}
} \\
\subfloat[Stochastic lift domain lift domain]{
\includegraphics[width=0.55\linewidth]{{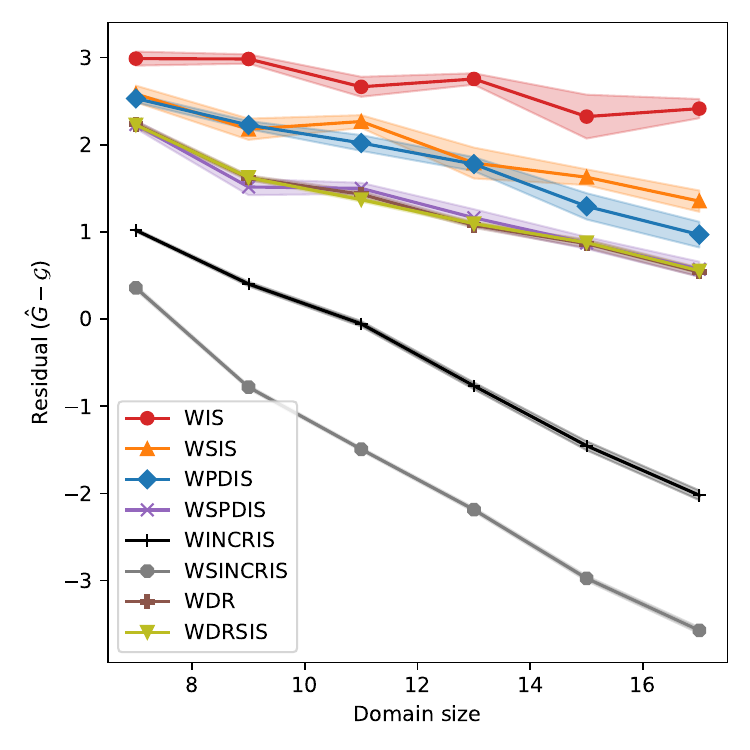}}
}
\caption{Residuals ($y$-axis), defined as $\hat{G} - \mathcal{G}$, as a function of the domain size ($x$-axis) in the lift domains. Estimates are based on 1,000 episodes. Residuals are represented by their mean $\pm$ standard error over 50 independent runs for the deterministic lift domain and over 200 independent runs for the stochastic domain. Note that SDRE and SSDRE are not included for improved visibility as their residuals are extremely large.}
\label{fig: lift} 
\end{figure}

\begin{figure}[htbp!]
\centerline{\includegraphics[width=0.55\linewidth]{{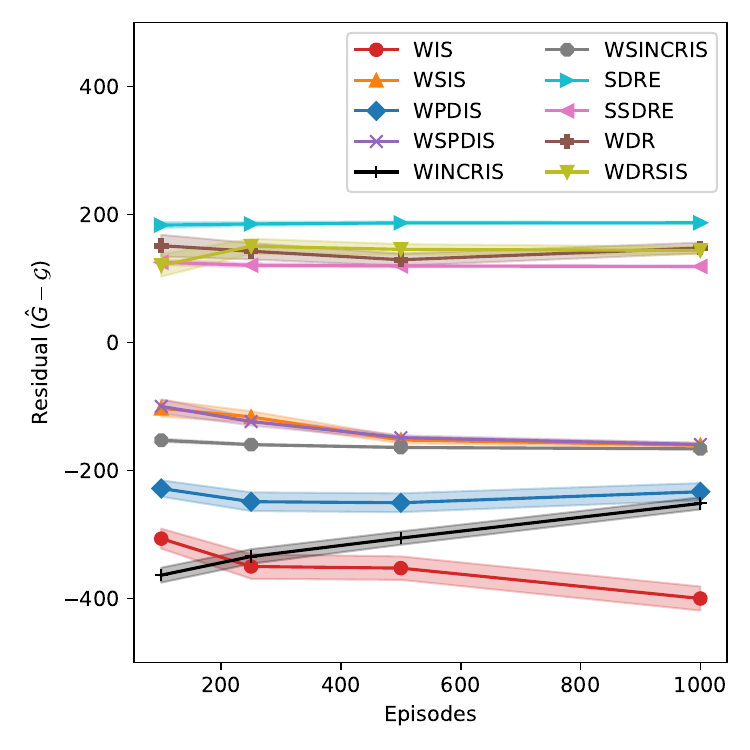}}}
\caption{Residuals ($y$-axis), defined as $\hat{G} - \mathcal{G}$, as a function of the number of episodes ($x$-axis) in the inventory management domain. Residuals are represented by their mean $\pm$ standard error over 50 independent runs.}
\label{fig: IM} 
\end{figure}

\begin{figure}[htbp]
\centerline{\includegraphics[width=0.55\linewidth]{{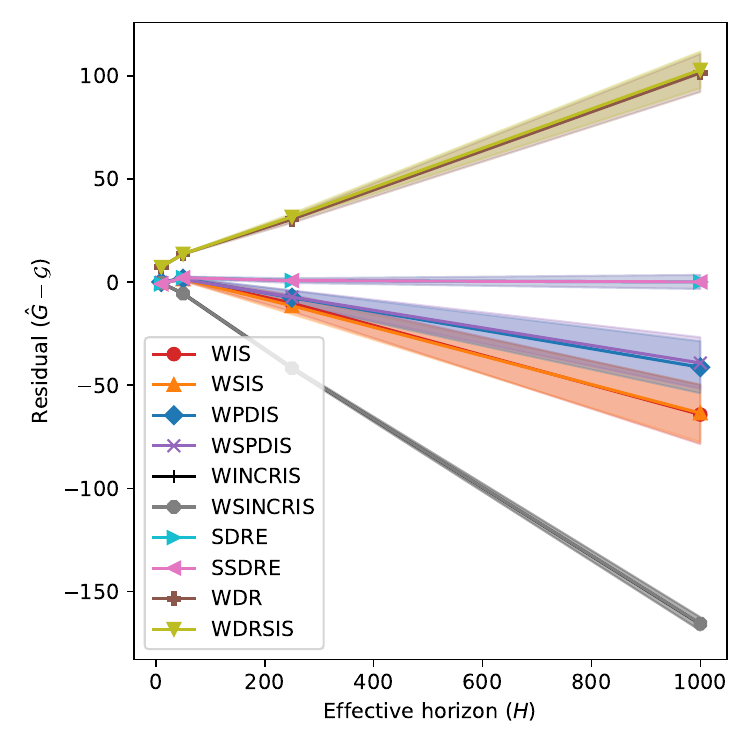}}}
\caption{\footnotesize{Residuals ($y$-axis), defined as $\hat{G}  - \mathcal{G}$, as a function of the effective horizon ($H$; $x$-axis) of the taxi domain. Residuals are represented by their mean $\pm$ standard error over 20 independent runs.}}\label{fig: taxi} 
\end{figure}

\end{document}